\documentclass[sigconf]{acmart}
\AtBeginDocument{%
  \providecommand\BibTeX{{%
    \normalfont B\kern-0.5em{\scshape i\kern-0.25em b}\kern-0.8em\TeX}}}

\setcopyright{acmcopyright}
\copyrightyear{2018}
\acmYear{2018}
\acmDOI{XXXXXXX.XXXXXXX}

\acmConference[ACM MM]{}{October 10--14,2022}{Lisbon, Portugal}
\acmPrice{15.00}
\acmISBN{978-1-4503-XXXX-X/18/06}

\acmSubmissionID{2739}

\usepackage{multirow}
\usepackage{colortbl}
\usepackage{float}

\renewcommand\footnotetextcopyrightpermission[1]{}

\begin{document}

% \title{RS-UNMVS: Unsupervised Multi-view Stereo via Image Rendering and Depth Synthesis}
\title{DS-MVSNet: Unsupervised Multi-view Stereo via Depth Synthesis}

%%
%% Author
\author{Jingliang Li}
\email{lijingliang20@mails.ucas.ac.cn}
\authornote{Both authors contributed equally to this research.}
\affiliation{%
  \institution{School of AI, University of Chinese Academy of Sciences}
  \country{}
}

\author{Zhengda Lu}
\email{luzhengda@ucas.ac.cn}
\authornotemark[1]
\affiliation{%
  \institution{School of AI, University of Chinese Academy of Sciences}
  \country{}
}

\author{Yiqun Wang}
\email{yiqun.wang@cqu.edu.cn}
\affiliation{%
  \institution{College of Computer Science, Chongqing University \& KAUST}
  \country{}
}

\author{Ying Wang}
\email{ywang@ucas.ac.cn}
\affiliation{%
  \institution{School of AI, University of Chinese Academy of Sciences}
  \country{}
}

\author{Jun Xiao}
\email{xiaojun@ucas.ac.cn}
\authornote{Corresponding author}
\affiliation{%
  \institution{School of AI, University of Chinese Academy of Sciences}
  \country{}
}

%% Define a more concise list of authors' names for this purpose.
\renewcommand{\shortauthors}{Jingliang Li et al.}

%%
%% Abstract

\begin{abstract}
In recent years, supervised or unsupervised learning-based MVS methods achieved excellent performance compared with traditional methods.
However, these methods only use the probability volume computed by cost volume regularization to predict reference depths and this manner cannot mine enough information from the probability volume.
Furthermore, the unsupervised methods usually try to use two-step or additional inputs for training which make the procedure more complicated.
In this paper, we propose the DS-MVSNet, an end-to-end unsupervised MVS structure with the source depths synthesis.
To mine the information in probability volume, we creatively synthesize the source depths by splattering the probability volume and depth hypotheses to source views.
Meanwhile, we propose the adaptive Gaussian sampling and improved adaptive bins sampling approach that improve the depths hypotheses accuracy.
On the other hand, we utilize the source depths to render the reference images and propose depth consistency loss and depth smoothness loss.
These can provide additional guidance according to photometric and geometric consistency in different views without additional inputs.
Finally, we conduct a series of experiments on the DTU dataset and Tanks $\&$ Temples dataset that demonstrate the efficiency and robustness of our DS-MVSNet compared with the state-of-the-art methods.
\end{abstract}


%   <concept>
%       <concept_id>10010147.10010178.10010224.10010245.10010254</concept_id>
%       <concept_desc>Computing methodologies~Reconstruction</concept_desc>
%       <concept_significance>500</concept_significance>
%       </concept>
%   <concept>
%       <concept_id>10010147.10010178.10010224.10010245.10010255</concept_id>
%       <concept_desc>Computing methodologies~Matching</concept_desc>
%       <concept_significance>300</concept_significance>
%       </concept>
%   <concept>
%       <concept_id>10010147.10010178.10010224.10010226.10010235</concept_id>
%       <concept_desc>Computing methodologies~Epipolar geometry</concept_desc>
%       <concept_significance>100</concept_significance>
%       </concept>
%   <concept>
%       <concept_id>10010147.10010178.10010224.10010226.10010239</concept_id>
%       <concept_desc>Computing methodologies~3D imaging</concept_desc>
%       <concept_significance>100</concept_significance>
%       </concept>
%  </ccs2012>
% \ccsdesc[500]{Computing methodologies~Reconstruction}
% \ccsdesc[300]{Computing methodologies~Matching}
% \ccsdesc[100]{Computing methodologies~Epipolar geometry}
% \ccsdesc[100]{Computing methodologies~3D imaging}

%%
%% Keywords.

%%
%% teaser image.
%   \includegraphics[width=\textwidth]{srcs/images/1_image.pdf}
%   \caption{Visualization of our synthetic source depths and rendered source images.}

\maketitle

\section{Introduction}

\begin{figure}
  \includegraphics[width=0.48\textwidth]{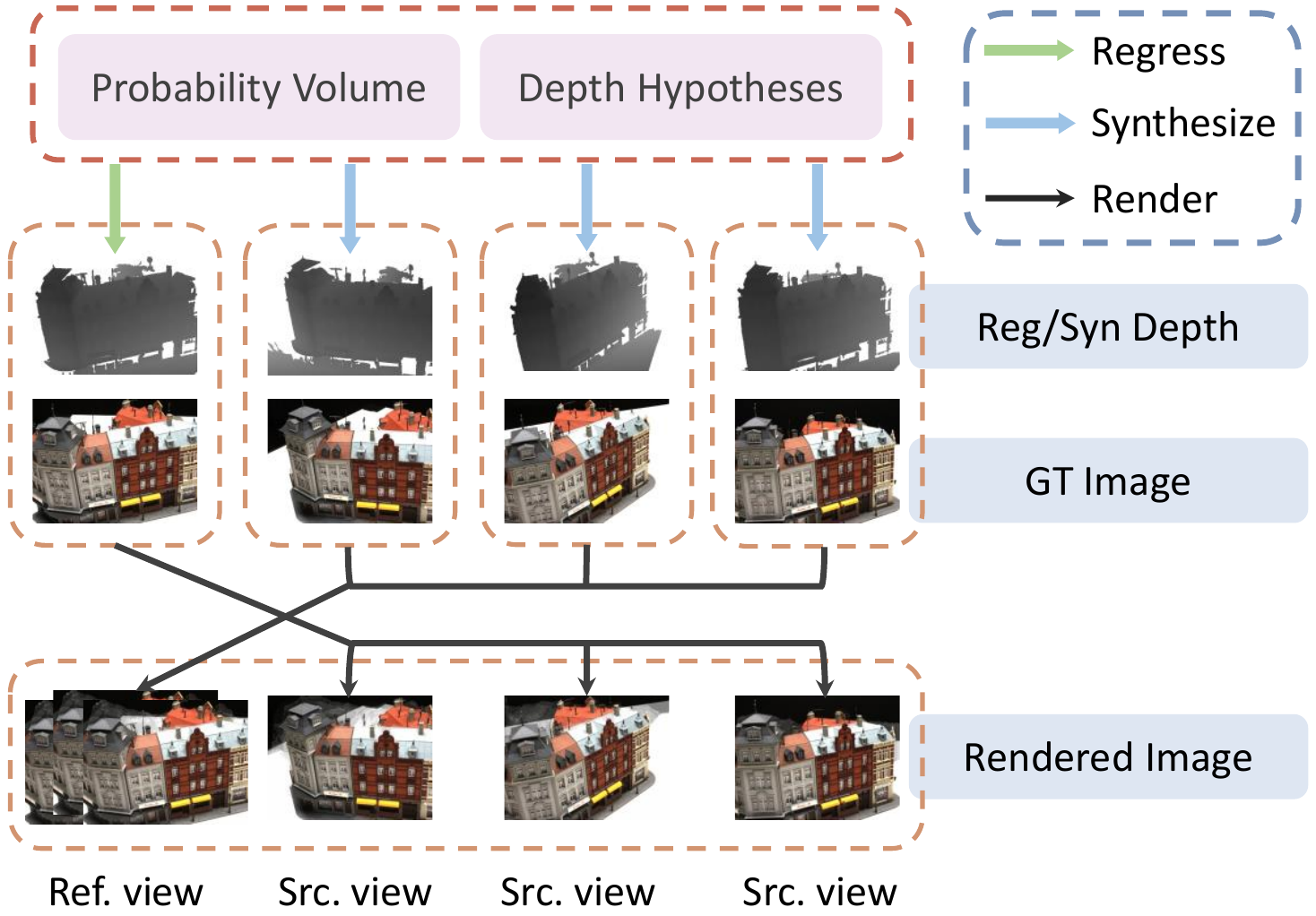}
  \caption{The procedure of depth synthesis and image rendering. In other methods, they regress the depth map of the reference view and use it to render the source view images.
Furthermore, we synthesize the depth map of each source view and render their corresponding reference image.}
  %The procedure of depth synthesis and image rendering. For reference view, we regress the depth map. In each source view, we calculate the depth map by synthesizing. Reference depth map and ground truth image use to render each source view image and each source depth map and ground truth image will render corresponding reference image.}
  \label{fig:img_01}
\end{figure}

\begin{figure*}
  \includegraphics[width=0.9\textwidth]{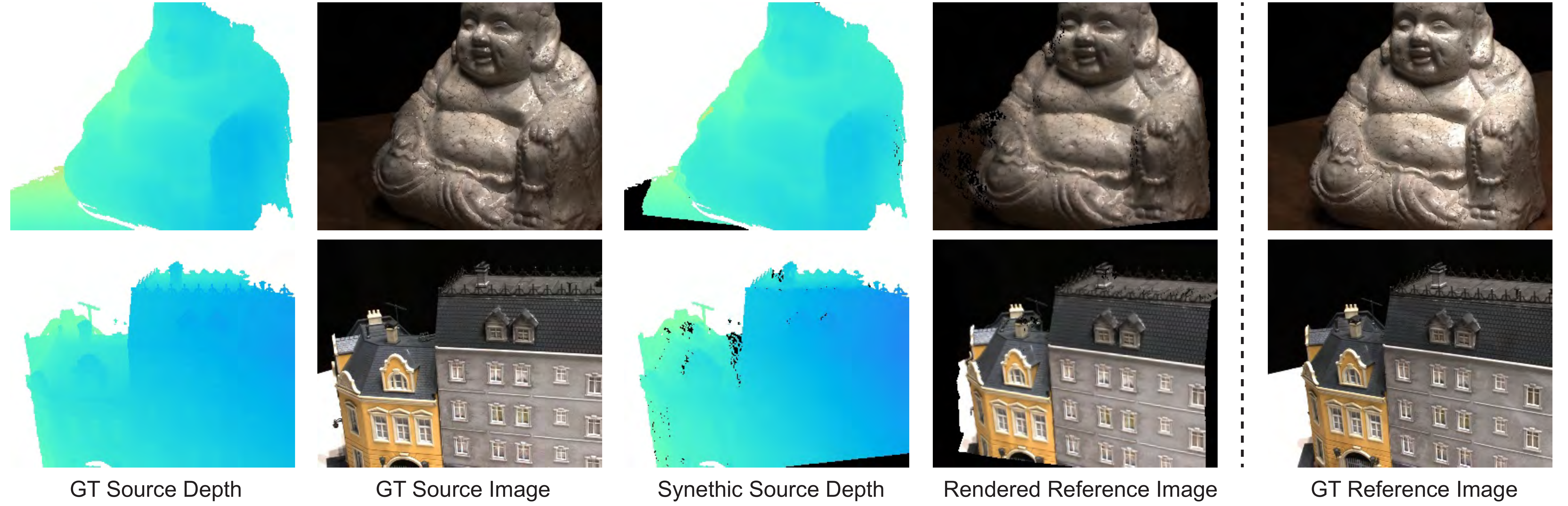}
  \caption{Visual illustrations of synthetic source depths and rendered reference images. The black region in synthetic depth represent the uncertainty region due to occlusion.}
  %Visual illustrations of synthetic source depths and rendered reference images. The rendered reference images are generate from source image, and the black region in synthetic depth and rendered image represent the uncertainty due to occlusion.}
  \label{fig:1_depth_img}
\end{figure*}

Multi-view stereo (MVS) is a fundamental topic in computer vision, which tackles the problem of dense 3D model reconstruction of a scene from a set of different-view images, with known camera parameters.
The applications in augmented reality, autonomous driving, and robotics have been studied for decades \cite{seitz2006comparison}.

While traditional MVS methods \cite{furukawa2009accurate,zheng2014patchmatch,galliani2015massively,schonberger2016pixelwise,xu2019multi} have achieved excellent performance, they are mainly limited by their hand-crafted features, which make dense matching intractable and thus lead to incomplete reconstructions. Inspired by the great success of Convolutional Neural Networks (CNNs), learning-based supervised MVS methods \cite{yao2018mvsnet, gu2020cascade, yang2020cost, wang2021patchmatchnet, wei2021aa} first extract image features and build cost volumes to capture geometric structures in different views. Then they generate the probability volume by cost volume regularization and regress reference depth.
Although the results of these methods are significantly better than traditional methods, their success relies heavily on ground-truth data, which is not readily available.

Currently, more and more attention has focused on unsupervised MVS methods that do not require ground-truth depth as input.
UnSup-MVS \cite{khot2019learning}, M\^{}3VSNet \cite{huang2021m3vsnet} proposed a robust photometric consistency loss to minimize the difference of the 3D point in different views and train their network.
Furthermore, \cite{xu2021digging,yang2021self} employed a two steps framework to apply the advantage of the complementary information between multi-views.
They first generate the initial pseudo depths by unsupervised learning and then use them as the supervised depth to train again. 
Although these methods achieve superior performance, they cannot be trained in an end-to-end manner.
Furthermore, some methods utilize additional inputs besides the multi-images, such as pre-processed optical flow \cite{xu2021digging}, augmented data which infer two times\cite{xu2021self} in training, and the pre-trained image semantic segmentation backbone \cite{xu2021self}.
These methods make the training more complicated and need more additional information.

Generally, the above two categories of methods both train the network by extracting supervised information from the images or other additional inputs.
By contrast, we mine effective supervision information by digging into the current backbone.
Current supervised or unsupervised methods predict reference depths by using the probability volume as shown in the left column in Fig. \ref{fig:img_01}.
However, the probability volume aggregates the similarity information between different view images by warping the source features to reference view. 
Therefore, we utilize the inherent information in the probability volume to synthesize the source depths, see Fig. \ref{fig:img_01}.
Finally, we conduct accurate synthetic source depths and realistic rendered reference images as illustrated in Fig. \ref{fig:1_depth_img}, and these images are valid to be the supervision information in training.

To this end, we propose the DS-MVSNet, a simplified end-to-end unsupervised MVS structure with the source depths synthesis.
Specifically, we conduct the synthetic source depths by splattering the reference probability volume and depth hypotheses to source views, which can mine the inherent information in the probability volume.
In this procedure, we propose the adaptive Gaussian sampling and improved adaptive bins sampling approach to improve the accuracy of the predicted reference depth and synthetic source depths.
On the other hand, we utilize the source depths to render the reference images and then help to train our network with photometric consistency loss on both reference and source views.
Furthermore, we propose depth consistency loss to minimize the depth difference of pixels belonging to the same 3D point in different views.
Finally, we propose depth smoothness loss with the reference image smoothness module, which can reduce the effect of shadows.

In summary, our contributions are the following:

\begin{itemize}
    \item We creatively synthesize the source depth by splattering probability volume and depth hypotheses, and propose the adaptive Gaussian sampling and improved adaptive bins sampling approach to improving the depths hypotheses efficacy.
    \item We propose depth consistency and depth smoothness loss and render reference images through source depths, which can provide additional guidance based on the geometric and photometric consistency in different views and ground truth images.
	\item We propose a simplified unsupervised MVS structure, DS-MVSNet, which is trained in an end-to-end manner and performs the dense reconstruction experiments on MVS datasets and achieves the best overall performance.
\end{itemize}
\section{Related Work} 

We review the most related works in the literature from three aspects, i.e. traditional MVS, supervised learning-based MVS and unsupervised learning-based MVS.

\begin{figure*}[th]
  \includegraphics[width=0.90\textwidth]{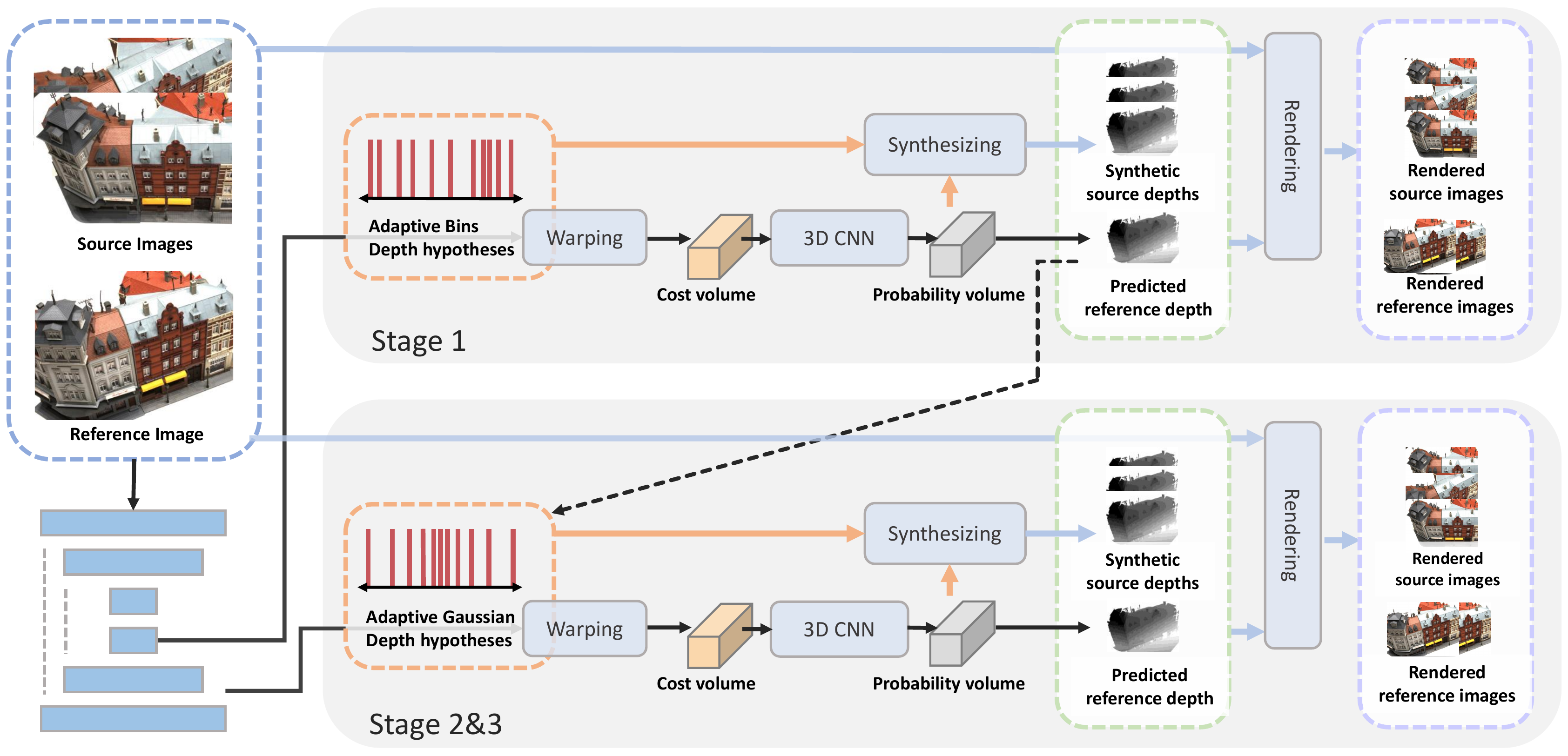}
  \caption{The framework of our DS-MVSNet. We construct the 3D cost volume based on our adaptive sampling depth hypotheses in each stage. Then we regularize it by 3D CNNs and get a probability volume to regress reference depth and synthesize source depths. Finally, we render each view image according to the depths and ground truth images.}
  \Description{Struct of our network.}
  \label{fig:3_network}
\end{figure*}

\subsection{Traditonal MVS}

Multi-view stereo has been extensively studied for decades. Before the deep learning era, many traditional methods have been proposed in this field.
According to output scene representations, there are mainly four types of reconstruction methods, such as volumetric representation \cite{goesele2006multi,kutulakos2000theory,seitz1999photorealistic}, point cloud \cite{furukawa2009accurate,lhuillier2005quasi}, mesh \cite{kazhdan2006poisson,curless1996volumetric,fua1995object} and depth map \cite{xu2019multi,schonberger2016structure,schonberger2016pixelwise,galliani2015massively,campbell2008using}.
In particular, volumetric based methods usually use two steps to reconstruct the scene, they first discretize 3D space into regular grids and then decide whether a voxel belongs to the near surface. However, space discretization is memory intensive and is not scalable to large-scale scenarios.
Point based methods usually start from a sparse set of matched key points and gradually densify the reconstruction using propagation strategies, which limits the capacity of parallel points processing.
Mesh based methods try to directly reconstruct surface meshes. These reconstructed surfaces often look smoother, and often lack high-frequency details.
In contrast, depth-based methods have shown more flexibility in reconstruction. It reduces the MVS reconstruction into relatively small problems of per-view depth map estimation and can be further fused to point cloud \cite{campbell2008using} or the volumetric reconstructions \cite{bhat2021adabins}.
Although stable results can be achieved, the traditional MVS pipeline mainly relies on hand-crafted similarity metrics, which make dense matching intractable and thus lead to incomplete reconstructions.

\subsection{Supervised learning-based MVS}
Recent supervised learning-based multi-view networks have shown great potential with better reconstruction accuracy and completeness. In particular, 3D cost volumes are widely used in recent state-of-the-art methods.
As a representative work, MVSNet \cite{yao2018mvsnet} constructs a 3D cost volume from features of different views via differentiable homography and regularizes the cost volume with 3D CNN for depth regression.
However, the memory consumption for 3D cost volume grows cubically. To reduce the huge memory consumption, some variants of MVSNet have been proposed recently, e.g. recurrent methods and coarse-to-fine methods.
Recurrent MVSNet architectures \cite{yao2019recurrent,yan2020dense,wei2021aa} sequentially regularize cost maps along the depth dimension with recurrent networks replacing the regularization network with cubic 3D convolutions. Recurrent based methods suffer from great time consumption in exchange for low memory cost. In contrast, coarse-to-fine methods \cite{yi2020pyramid,gu2020cascade,yang2020cost,wang2021patchmatchnet} preserve regularization with cubic 3D CNN. They first predict a low resolution depth map with a large depth interval and iteratively incremental resolution with decreasing depth range, which has high efficiency in time and memory.
Meanwhile, another line of research explores the performance of the pipeline. such as \cite{xu2020learning,cheng2020deep} proposed effective depth sampling approach, \cite{luo2020attention,ma2021epp,yu2021attention} used attention strategy improving the representation ability of the feature for image information, \cite{zhang2020visibility} considered the visibility around different views using a uncertainty part.

\subsection{Unsupervised learning-based MVS}
To get rid of the dependence on ground truth and improve the generalization ability of the model, unsupervised learning-based methods have emerged as an alternative based on the fundamental assumption of photometric consistency and achieved competitive results.
For instance, UnSup-MVS \cite{khot2019learning}, the first end-to-end learning-based unsupervised MVS network, minimizes the difference of photometric between the real and warped images according to predicted reference depth.
M\^{}3VSNet \cite{huang2021m3vsnet} enforce the consistency between surface normal and depth map to regularize the MVS pipeline.
Different views of a scene can be constrained to each other to improve depth accuracy. Self-supervised CVP-MVSNet \cite{yang2021self}, U-MVS \cite{xu2021digging} proposed a two steps framework. They first generate initial pseudo labels and then refine the pseudo labels by iteration. However, these methods cannot be trained in an end-to-end manner.
JDACS \cite{xu2021self} proposed an end-to-end network, supervised by photometric consistency, segmentation map and augmentation data. However, it requires a pretrained feature extraction backbone for segmentation and inferring two times due to data augmentation.

\section{Method}

For the task of multi-view images reconstruction, we propose the DS-MVSNet, a novel end-to-end structure of the unsupervised learning-based MVS method. The overall framework is depicted in Fig. \ref{fig:3_network}.
In this section, we first describe the overall architecture of the DS-MVSNet. Then we describe the details of the source depth synthesis module in Sec. \ref{subsec:depth_synthesis}, which includes probability volume splattering and adaptive depth hypotheses sampling.
Finally, we introduce the unsupervised training loss in Sec. \ref{subsec:training_loss}.

\subsection{Network Architecture} \label{subsec:supervised_pipeline}

Recent coarse-to-fine method\cite{gu2020cascade} mainly inherited from MVSNet \cite{yao2018mvsnet} and contains three coarse-to-fine stages with four common procedures including feature extraction, cost volume construction, cost regularization, and depth regression.
Meanwhile, the CasMVSNet \cite{gu2020cascade} is friendly to GPU memory and can infer high-resolution outputs with more accurate predicted depth maps.
Thus, we utilize it as the backbone of our unsupervised framework.
As shown in Fig. \ref{fig:3_network}, our DS-MVSNet contains three stages with incremental resolution generated by repeating the above four procedures. 
Moreover, the depth hypotheses with decreasing depth range are sampled based on the depth map predicted by the previous stage.

To start with, given the input of $N$ images of size $H \times W$, we use $I_1$ and ${\{\textbf{I}_i\}}_{i=2}^{N}$ to denote the reference and source images which take from different viewpoints.
First, we extract multi-scale features $\{\textbf{F}_{i,k}\}_{k=1}^{3}$ of all images through a small FPN \cite{lin2017feature} with shared weights and the resolutions of these corresponding features are $H/4 \times W/4$, $H/2 \times W/2$ and $H \times W$ respectively.

After that, we construct the 3D cost volumes, which is a critical step for learning-based MVS \cite{yao2018mvsnet}.
For this purpose, we first conduct the adaptive depth hypotheses sampling to get $M$ depth hypotheses $\{d_{j, k}\}_{j=1}^{M}$, sampled from the whole known depth range. 
While $d_{1,k}$ represents the minimum depth and $d_{M, k}$ represents the maximum depth at the $k$-th stage, we ignore the stage index $k$ to simplify the description in the following formulation.
With these hypotheses, we construct a set of feature volumes ${\{\textbf{V}_i\}}_{i=1}^{N}$ by differentiable warping the 2D source image features to the reference camera views.
Meanwhile, the homography between the features of $i$-th source and reference image at depth $d$ is followed as::
\begin{equation}
    \textbf{H}_{i}(d) = d\textbf{K}_i\textbf{T}_i\textbf{T}_{1}^{-1}\textbf{K}_{1}^{-1}
    \label{eq:homography}
\end{equation}
where $\textbf{K}_1$, $\textbf{K}_i$ denote the intrinsic matrix of reference and $i$-th source image, and $\textbf{T}_1$, $\textbf{T}_i$ refer to extrinsic matrix of reference and $i$-th source image respectively.

In order to handle arbitrary number of input views, their multiple feature volumes ${\{\textbf{V}_i\}}_{i=1}^{N}$ need to be aggregated to one 3D cost volume $\textbf{V}$. Thus, we use the variance-based aggregation strategy same as CasMVSNet \cite{gu2020cascade}, which can be modeled as:
\begin{equation}
    \textbf{V} = \frac{1}{N} \sum_{i=1}^{N} {(\textbf{V}_i - \bar{\textbf{V} })}^2
    \label{eq:cost_aggregation}
\end{equation}
Where $\bar{\textbf{V}}$ denotes the average feature volume.

Next, we regularize the cost volume with a 3D U-Net \cite{ronneberger2015unet} network and produce a probability volume $\textbf{P}$ following a softmax-based function.
Specifically, the probability volume is the weight of different depth hypotheses and is used for reference depth regression.
Finally, the regressed depth in reference view at pixel $\textbf{p}_r$ is calculated as:
\begin{equation}
    \textbf{D}_{r}(\textbf{p}_r) = \sum_{j=1}^{M}d_j \cdot \textbf{P}(\textbf{p}_r, j)
    \label{eq:regression}
\end{equation}

\begin{figure}[t]
  \includegraphics[width=0.40\textwidth]{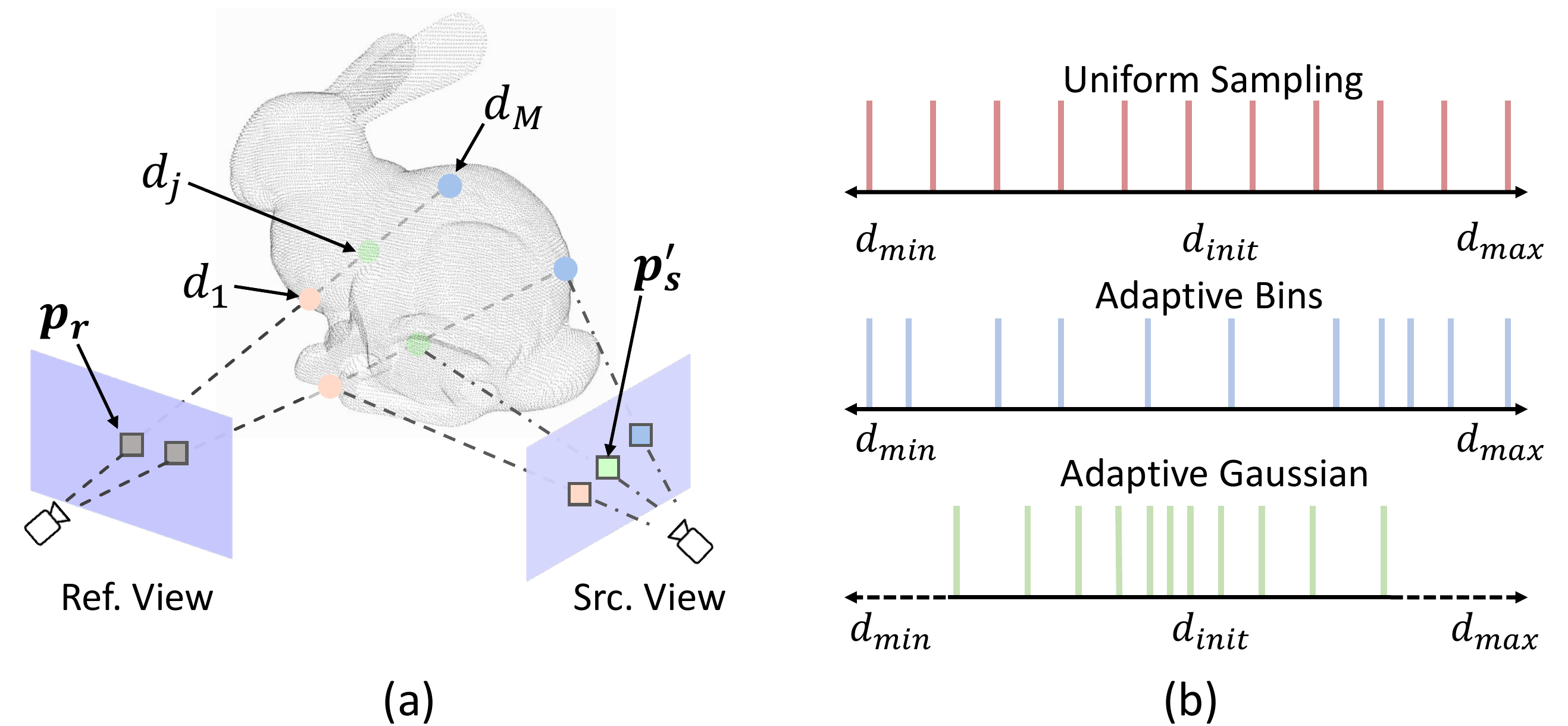}
  \caption{Pixel Projection and Sampling.
  (a) Visualization of the point correspondence between reference and source images at different depth hypotheses.
  (b) Comparison with different depth hypotheses sampling.
  Uniform sampling is usually used in previous works.
  We use the improved adaptive bins sampling in the coarse stage and the adaptive Gaussian sampling in the fine stage.
  The $d_init$ represents the pixel depth value predicted by the previous stage.
  }
  \label{fig:3_homography}
\end{figure}

\subsection{Source depths prediction via Synthesizing} \label{subsec:depth_synthesis}

As aforementioned, the probability volume aggregates the similarity and difference information between reference and source images.
However, both supervised and unsupervised MVS networks only used it to infer the reference depth and cannot fully exploit its information.
To solve this shortcoming, we propose a reasonable module for synthesizing source depths from the probability volume, which is comprised of the probability volume splattering and adaptive depth hypotheses sampling.

\begin{figure}[t]
  \includegraphics[width=0.40\textwidth]{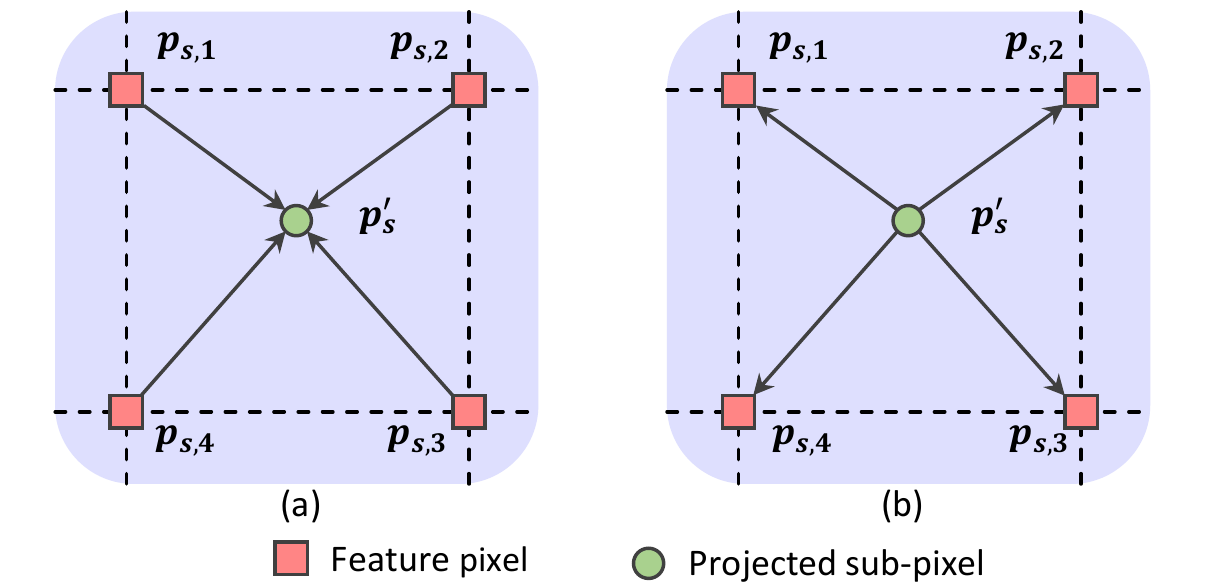}
  \caption{Comparison of the projected sub-pixel interpolation and splattering. 
(a) Interpolation of four feature pixels in the source view.
(b) Splattering to four feature pixels in the source view.}
  %Comparison of interpolation and splatting. (a) Interpolation of one projected sub-pixel for four feature pixel in source view, used in other methods. (b) Splatting of one projected sub-pixel to four feature pixels in source view, used in our method.}
  \label{fig:3_splatting}
\end{figure}

\paragraph{\textbf{Probability volume Splattering}}
As shown in Fig. \ref{fig:3_homography} (a), we propose the probability volume splattering to project the pixel in reference view at different depth hypotheses to the source views.
Let $\textbf{p}_r$ be the pixel in reference view, $\{d_j\}_{j=1}^{M}$ and $\{w_j\}_{j=1}^{M}$ are the depth hypotheses and the corresponding probability values of $\textbf{p}_r$.
Then we can compute its projected position $\textbf{p}_s^{\prime}$ and projected depth $d_j^{\prime}$ by the camera parameters according to Eq. \ref{eq:homography}. The projected probability value in source view $w_j^{\prime}$ is the same as the reference view.

For more details, $\textbf{p}_s^{\prime}$ is a sub-pixel, located in four neighbor pixels of the source view, we splatter it to the four neighbor pixels $\{\textbf{p}_{s,i}\}_{i=1}^{4}$, which is an inverse process of bilinear interpolation, as illustrated in Fig. \ref{fig:3_splatting}.
For the source pixel $\textbf{p}_s$, let $\mathcal{D}$ and $\mathcal{W}$ represent the set of splattered depth hypotheses and corresponding probability values in the source view respectively.
However, many reference points may contribute to the same source pixel and some pixels may have no corresponding points due to the occlusion between different views and the discreteness of depth hypotheses.
Thus, we use a binary mask $\mathbf{M}_d$ to ignore such pixels. And we compute the synthetic depth of source pixel $\textbf{p}_s$ as:
\begin{equation}
    \textbf{D}_{s}(\textbf{p}_s)=\mathbf{M}_d' \cdot \frac{{\sum_{d',w'} d'\cdot w'}}
    {{\sum w'} \enspace + \enspace \epsilon}, \quad {d'\in\mathcal{D},{w'\in\mathcal{W}}}
    \label{eq:synthesizing_depth}
\end{equation}

\begin{equation}
    \text{ where } \textbf{M}_d' = \begin{cases} 1 & \text{ if } \sum w' >\tau, {w'\in\mathcal{W}} \\ 0 & \text{ otherwise} \end{cases}
    \label{eq:synthesizing_mask}
\end{equation}
The small $\epsilon$ in the denominator ensures numerical stability for source pixels that correspond to no splatted points. And $\tau$ indicates the probability threshold.

\paragraph{\textbf{Adaptive depth hypotheses sampling}}
Depth hypotheses play a vital role in reference depth prediction and source depths synthesis.
Recent learning-based methods utilize uniform sampling to get the depth hypotheses which are discrete and lots of them are far from real depth.
In general, these faraway hypotheses have a small probability in the probability volume and then have less contribution for the pixel matching.
Furthermore, their matched pixels may have an inaccurate depth and be even ignored with $M(\textbf{p}_s):=0$, according to Eq. \ref{eq:synthesizing_depth} and Eq. \ref{eq:synthesizing_mask}.
Therefore, we generate more effective depth hypotheses based on improved adaptive bins sampling and our proposed adaptive Gaussian sampling, as shown in Fig. \ref{fig:3_homography}(b).

In the coarse stage ($k=1$), we implement the improved adaptive bins sampling to generate $M$ depth hypotheses for reference view.
Instead of the depth hypotheses being the same for all pixels in AdaBins \cite{bhat2021adabins}, our improved adaptive bins sampling can generate different hypotheses for each pixel.
First, we obtain $M$ bin-widths $\left \{ b_j  \right \}_{j=1}^{M}$ for each pixel using a CNNs block.
The input of the CNNs is the reference feature $\textbf{F}$ in the coarse stage, and the output has $M$ channels at each pixel.
Finally, the depth hypotheses $d_j$ of $\textbf{p}$ are defined as:
\begin{equation}
    d_j = d_{min} + (d_{max} - d_{min})(b_i/2+\sum_{i=1}^{j-1}b_j)
    \label{eq:adabins}
\end{equation}

Furthermore, we propose an adaptive Gaussian sampling used in the fine stage ($k=2,3$) to generate more effective hypotheses.
As shown in Fig. \ref{fig:3_network}, the depth range is decreased based on the predicted reference depth map in the previous stage.
For each pixel, we first define a Gaussian distribution whose mean is zero and the variance is set to one.
Since the initial depth of the hard samples may be inaccurate in the occlusion or texture-less area, we compute an entropy map $\textbf{E}$ to measure the uncertainty, following as:
\begin{equation}
    \textbf{E}(\textbf{p}_r)= - \sum_{m=1}^{M} \textbf{P}(\textbf{p}_r,j) \log (\textbf{P}(\textbf{p}_r,j))
    \label{eq:entropy}
\end{equation}

According to entropy map, we adaptively adjust the interval of Gaussian distribution $[-1 - \textbf{E}(\textbf{p}_r), 1 + \textbf{E}(\textbf{p}_r)]$.
Then, we calculate the bin-widths for the cumulative distribution function of the Gaussian distribution interval by uniform sampling and use the softmax layer to normalize.
Meanwhile, we use Eq. \ref{eq:adabins} to generate adaptive depth hypotheses same as the coarse stage.
Finally, the depth hypotheses are dense near the initial depth value and sparse away from the center as shown in Fig. \ref{fig:3_homography} (b), which leads to highly efficient spatial partitioning.

\subsection{Training loss} \label{subsec:training_loss}

Previous works only pay attention to predicted reference depth by warping reference images. We further utilize the rendered source images and synthetic source depths to optimize our model.
%The innovations of our unsupervised loss compared to previous methods are: a novel depth consistency loss to ensure the depth coherence in reference-source depth, robust depth smoothness loss using reference image smoothness module, and additional photometric consistency loss for source views.

In general, our unsupervised loss contains four components, photometric consistency loss $\mathcal{L}_{pc}$ for reference and source views, structure similarity loss $\mathcal{L}_{ssim}$ in reference and source views, depth consistency loss $\mathcal{L}_{dc}$ and depth smoothness loss $\mathcal{L}_{ds}$, at each scale $k$.
The total loss forms as the sum of $\mathcal{L}=\sum_{k=1}^{3}\mathcal{L}_{k}$. Our loss module at each scale is computed as:

\begin{equation}
    \mathcal{L}_{k}={\lambda}_{1}\mathcal{L}_{pc}+{\lambda}_{2}\mathcal{L}_{ssim}+{\lambda}_{3}\mathcal{L}_{ds}+{\lambda}_{4}\mathcal{L}_{dc}
    \label{eq:total_loss}
\end{equation}
Where ${\lambda}_i$ is the hyperparameter to weight the respective loss term.

Predicted reference depth and synthetic source depths have been mentioned in Sec. \ref{subsec:supervised_pipeline}, \ref{subsec:depth_synthesis}.
Thus, we next present the reference and source image rendering, the reference image smoothness module, and each component of our loss respectively.
\begin{figure}[t]
  \includegraphics[width=0.40\textwidth]{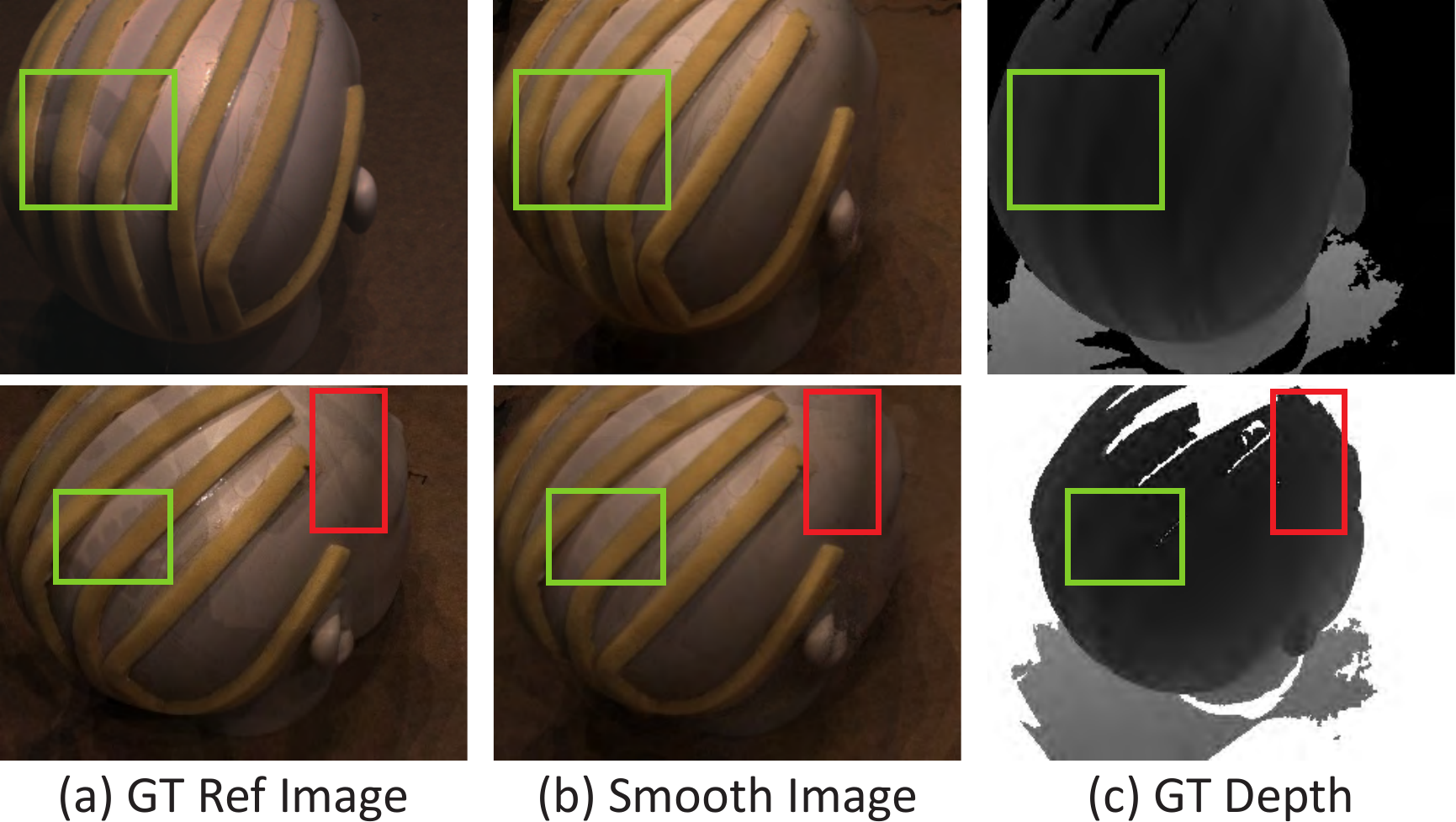}
  \caption{Reference image smoothness. In green and red rectangle, the depth is continuous, but the ground truth reference image is affected by shadows. Our smoothness reference image is more clean compared with ground truth.}
  \label{fig:3_smoothness}
\end{figure}

\paragraph{\textbf{Reference and source images rendering}}
For a pair of reference and source images $(\textbf{I}_1, \textbf{I}_i)$, we first compute the rendered source image $\hat{\textbf{I}}_i^1$ from the real reference image $\textbf{I}_1$ with the intrinsic and extrinsic camera parameters $([\textbf{K}_1, \textbf{T}_1], [\textbf{K}_i, \textbf{T}_i])$ and predicted reference depth $\textbf{D}_r$.
Meanwhile, we also use the homography Eq. \ref{eq:homography} to find corresponding pixels in the source view, and the rendered source image $\hat{\textbf{I}}_i^1$ can be calculated via the reference pixel texture splattering such as the \textit{Probability volume Splattering} in Sec. \ref{subsec:depth_synthesis}.
In addition to the rendered image $\hat{\textbf{I}}_i^1$, we generate a binary mask $\textbf{M}_{i}^{1}$ to mask out invalid pixels that are projected outside the source image bounds.

To this end, each source image have a corresponding rendered image in $\{\hat{\textbf{I}}_i^1\}_{i=2}^{N}$ and we generate $N-1$ rendered reference images $\{\hat{\textbf{I}}_1^i\}_{i=2}^{N}$ by source images, synthetic source depths and corresponding camera matrix respectively.

\paragraph{\textbf{Reference image smoothness module}}
%As shown in Fig. \ref{fig:3_smoothness} (a), the reference image with obvious shadows will lead to incorrect guidance in depth smoothness loss. 
As shown in Fig. \ref{fig:3_smoothness} (a), the shadows in different view images are different due to the lighting effects, which lead to incorrect guidance in the depth prediction. Furthermore, the shadow boundaries in these images are not consistent with the smooth boundaries in the depth map.
Thus, we propose the reference image smoothness module to attenuate the influence of shadows in the reference image.
Benefit from reference images rendering, we generate a smooth reference image $\tilde{\textbf{I}}_1$ via the rendered reference images $\{\hat{\textbf{I}}_1^i\}_{i=1}^{N}$:

\begin{equation}
    \tilde{\textbf{I}}_1(\textbf{p}_r) =
                0.5 \cdot \textbf{I}_1
              + 0.5 \cdot \frac{\sum_{i=2}^{N} \textbf{I}_1^i(\textbf{p}_r) \cdot \textbf{M}_1^i(\textbf{p}_r)}
                               {\sum_{i=2}^{N} \textbf{M}_1^i(\textbf{p}_r)}
    \label{eq:ref_smooth}
\end{equation}

As shown in Fig. \ref{fig:3_smoothness} (b), Our smoothed reference image becomes more continuous and more suitable as a guide for depth smoothness.

\paragraph{\textbf{Appearance Match Cost}}
The core of appearance match aims at minimizing the difference between the rendered image and the real image.
Inspired by \cite{khot2019learning, dai2019mvs2, huang2021m3vsnet}, we use a combination of a photometric consistency loss $\mathcal{L}_{pc}$ and single scale $SSIM$ term $\mathcal{L}_{ssim}$ \cite{wang2004image} as our appearance match cost for both reference and source views.

\begin{equation}
    \mathcal{L}_{ssim} = \sum_{i=2}^{N}
                        [1 - SSIM(\hat{\mathbf{I}}_i^1, \mathbf{I}_i)] + 
                        [1 - SSIM(\hat{\mathbf{I}}_1^i, \mathbf{1}_i)]
    \label{eq:loss_ssim}
\end{equation}

\begin{align}
  \begin{split}
  \mathcal{L}_{pc} &= \sum_{i=2}^{N}
                    \left \| (\nabla \hat{\mathbf{I}}_i^1 - \nabla\mathbf{I}_i) \odot \mathbf{M_i^1} \right \|
                    + (\left \| \hat{\mathbf{I}}_i^1 - \mathbf{I}_i\right \| \odot \mathbf{M_i^1} ) \\
                   &+ \left \| (\nabla \hat{\mathbf{I}}_1^i - \nabla\mathbf{I}_1) \odot \mathbf{M_1^i} \right \|
                    + (\left \| \hat{\mathbf{I}}_1^i - \mathbf{I}_1\right \| \odot \mathbf{M_1^i} )
  \end{split}
  \label{eq:loss_photo}
\end{align}

\paragraph{\textbf{Depth Consistency loss}}
Benefit from our novel source depths synthesis branch, we propose the depth consistency loss with L1 reference-source depth consistency penalty to ensure the depth coherence in reference-source depth.
First, we projected the predict reference depth to the source views $\{\tilde{\textbf{D}}_i^1\}_{i=2}^N$, according to Eq. \ref{eq:homography}. Then we attempt to make the projected depth map $\tilde{\textbf{D}}_i^1$ be equal to the synthetic source depth map $\textbf{D}_i$,
\begin{equation}
    \mathcal{L}_{dc}=\frac{1}{N-1} \sum_{i=2}^{N} \left | \tilde{\textbf{D}}_i^1 - \textbf{D}_i \right |
\end{equation}

\paragraph{\textbf{Depth Smoothness loss}}
Similar to \cite{mahjourian2018unsupervised}, to encourage smoother gradient changes, we propose the depth smoothness loss which is an edge-ware smoothness term based on smoothness reference image $\tilde{\textbf{I}}_1$ and is employed as:
\begin{equation}
    \mathcal{L}_{ds}=e^{\left|{\nabla}^x \tilde{\mathbf{I}}_r\right|} \cdot {\nabla}^x \mathbf{D}_{r}
                    +e^{\left|{\nabla}^y \tilde{\mathbf{I}}_r\right|} \cdot {\nabla}^y \mathbf{D}_{r}
    \label{eq:depth_smooth}
\end{equation}
\section{Experiments}

\begin{figure*}[t]
  \includegraphics[width=0.90\textwidth]{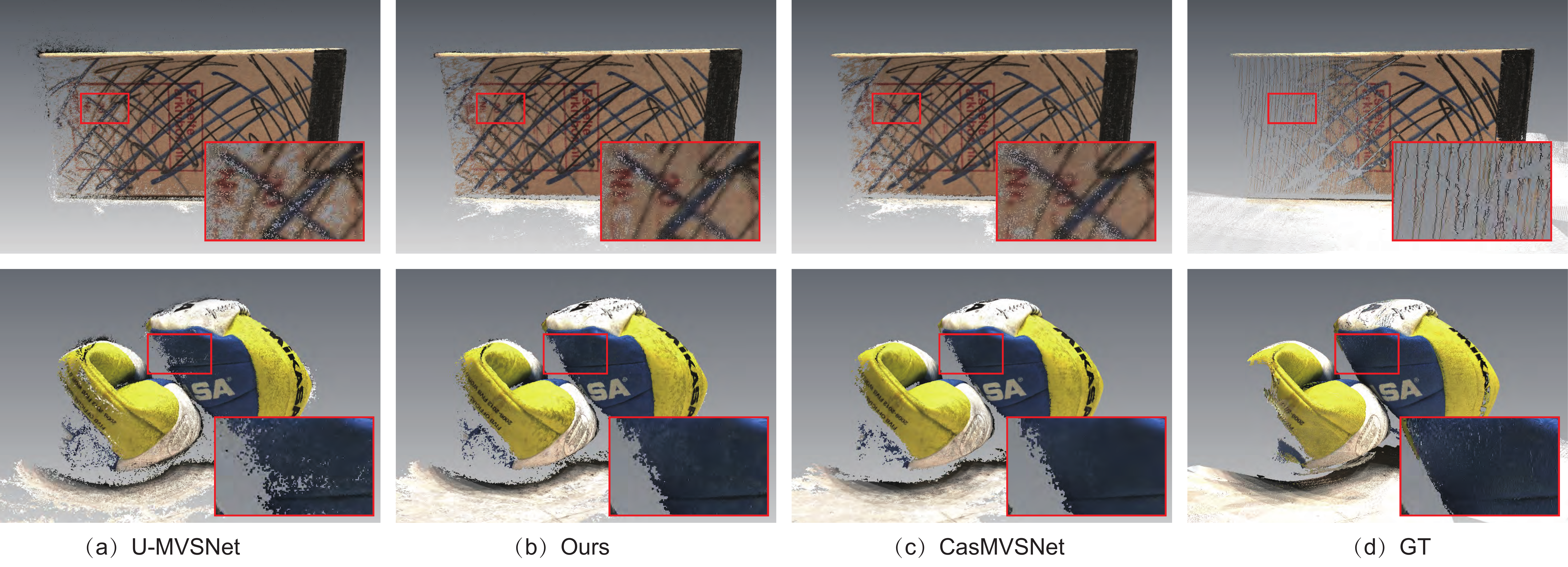}
  \caption{Comparison of reconstructed results with unsupervised method (U-MVSNet), supervised method (CasMVSNet), and ground truth on scan 10 and scan 62 in DTU dataset.
  }
  \label{fig:4_results_dtu}
\end{figure*}

This section demonstrates the performance of our DS-MVSNet with comprehensive experiments and verifies the effectiveness of proposed modules through ablation studies. We first introduce the datasets and implementation details and then analyze our results.

\subsection{Datasets}

We evaluate our model on DTU \cite{aanaes2016large} and Tanks and Temples \cite{knapitsch2017tanks} benchmark. The DTU dataset is a large-scale indoor multi-view stereo dataset collected under well-controlled laboratory conditions with a fixed camera trajectory. It consists of 124 different scenes and each scene has 49 images or 64 images under 7 different light conditions. We adopt the same training, validation, and evaluation split as defined in \cite{yao2018mvsnet}.
Tanks and Temples are collected in a more complex realistic environment, contain both indoor and outdoor scenes, and are split into intermediate and advanced.  We evaluate the generalization ability of our proposed method on the intermediate set, which consists of eight scenes for comparison.

\subsection{Implementation Details}

The proposed DS-MVSNet is trained on the DTU dataset. We first scale the DTU images to $640 \times 512$ resolution, and then generate coarse ground truth maps following the previous MVS methods \cite{yao2018mvsnet,gu2020cascade}.
In this paper, DS-MVSNet is implemented in 3 stages with $1/4$, $1/2$, and 1 of original input image resolution respectively. We separately sampled 48, 32, and 8 depth intervals from the first stage to the third stage. And all of the networks are trained with $N=3$.
In probability volume splattering, the probability threshold $\tau$ is set 0.001. The weight hyperparameters ${\lambda}_1,{\lambda}_2,{\lambda}_3,{\lambda}_4$ of the loss term are set to 12.0, 6.0, 0.05 and 0.01, respectively.
Finally, we implemented our network using Pytorch, utilize a batch size of 8, and trained on Adam optimizer for 8 epochs with a learning rate of 0.0001, and halved iteratively at the 6th, 7th, and 8th epochs.
All experiments were conducted on a server computer equipped with four NVIDIA TITAN RTX 24GB graphics cards.

\subsection{Evaluation on DTU dataset}

\begin{table}[t]
  \caption{Quantitative reconstruction results on DTU. Lower means better and the best result is bold. The sections are partitioned into supervised and end-to-end unsupervised, respectively. All methods do not use data augmentation. The results other than ours are from previously published literature.
  }
  \centering
  \begin{tabular}{
  p{1.8cm}p{2.6cm}p{0.8cm}<{\centering}p{0.8cm}<{\centering}p{0.8cm}<{\centering}
}
  \toprule
  & Method & Acc. & Comp. & Overall \\
  \midrule
%   \multirow{4}{*}{Traditional} & Furu \cite{furukawa2009accurate} & 0.613 & 0.941 & 0.777 \\
%   & Tola \cite{tola2012efficient} & 0.342 & 1.190 & 0.766 \\
%   & Camp \cite{campbell2008using} & 0.835 & 0.554 & 0.695 \\
%   & Gipuma \cite{galliani2015massively} & 0.283 & 0.873 & 0.578 \\
%   \midrule
  \multirow{7}{*}{Supervised}
  & MVSNet \cite{yao2018mvsnet} & 0.396 & 0.527 & 0.462 \\
%   & CIDER \cite{xu2020learning} & 0.417 & 0.437 & 0.427 \\
  & CasMVSNet \cite{gu2020cascade} & 0.325 & 0.385 & 0.355 \\
  & UCS-Net \cite{cheng2020deep} & 0.338 & 0.349 & \textbf{0.344} \\
  & CVP-MVSNet \cite{yang2020cost} & \textbf{0.296} & 0.406 & 0.351 \\
  & PatchmatchNet \cite{wang2021patchmatchnet} & 0.427 & \textbf{0.277} & 0.352 \\
  & AA-RMVSNet \cite{wei2021aa} & 0.376 & 0.339 & 0.357 \\
  & EPP-MVSNet \cite{ma2021epp} & 0.413 & 0.296 & 0.355 \\
  \midrule
%   \multirow{2}{*}{Two-Step}
%   & Yang et al. \cite{yang2021self} & \textbf{0.308} & 0.418 & 0.363 \\
%   & U-MVSNet-MS \cite{xu2021digging} & 0.354 & \textbf{0.353} & \textbf{0.354} \\
%   \midrule
  \multirow{7}{*}{Unsupervised}
  & UnSup-MVSNet \cite{khot2019learning} & 0.881 & 1.073 & 0.977 \\
  & MVS2 \cite{dai2019mvs2} & 0.760 & 0.515 & 0.637 \\
  & M\^{}3VSNet \cite{huang2021m3vsnet} & 0.636 & 0.531 & 0.583 \\
%   & JDACS \cite{xu2021self} & \textbf{0.571} & \textbf{0.515} & \textbf{0.543} \\
  & JDACS-MS${}$ \cite{xu2021self} & 0.443 & 0.389 & 0.416 \\
  & U-MVSNet-MS${}$ \cite{xu2021digging} & 0.375 & 0.383 & 0.379 \\
  & Ours & \textbf{0.374} & \textbf{0.347} & \textbf{0.361} \\
  \bottomrule
\end{tabular}
  \label{tab:4_results_dtu}
\end{table}

\begin{table}[t]
  \caption{Comparison of GPU memory and runtime on DTU with our backbone. These are obtained by running the official code on a TITAN RTX graphics card, with one batchsize.
  }
  \centering
  \begin{tabular}{
  p{2.2cm}<{\centering}p{1.5cm}<{\centering}p{1.8cm}<{\centering}p{1.5cm}<{\centering}
}
  \toprule
  Methods & Img Size & Mem. (MB) & Runtime (s) \\
  \midrule
  CasMVSNet \cite{gu2020cascade} & $640 \times 512$ & 6965 & 1.703 \\
  Ours & $640 \times 512$ & 9233 & 2.064 \\
  \bottomrule
\end{tabular}
  \label{tab:4_effective_dtu}
\end{table}

In this section, we compare our proposed DS-MVSNet with the state-of-the-art methods including supervised and unsupervised methods on the DTU dataset. 
Note that, we use 7 neighboring views and set the image size to be $1152 \times 864$ in the test set.
The metric evaluates point clouds using official evaluation scripts of DTU \cite{aanaes2016large}. It compares the distance between ground-truth point clouds and the produced point clouds. The state-of-the-art comparison results are shown in Tab. \ref{tab:4_results_dtu}.
In the unsupervised sections, the first four methods are one-stage approaches, the others are coarse-to-fine approaches. All of the method are not use data augmentation.
As shown in Tab. \ref{tab:4_results_dtu}, DS-MVSNet achieves the best accuracy, completeness, and overall score among coarse-to-fine supervised methods. Especially for completeness, the metric is improved by 10\% compared with U-MVSNet-MS \cite{xu2021self}.
Fig. \ref{fig:4_results_dtu} shows some qualitative results compared with other methods. We can see that our model can generate more complete point clouds with finer details.

\begin{table*}[t]
  \caption{
    Quantitative results of F-scores (higher means better) on intermediate subsets of Tanks and Temples.
  }
  \centering
  \begin{tabular}{
    p{2.0cm}p{3.2cm}p{0.9cm}<{\centering}
    p{0.9cm}<{\centering}p{0.9cm}<{\centering}p{0.9cm}<{\centering}p{0.9cm}<{\centering}
    p{0.9cm}<{\centering}p{0.9cm}<{\centering}p{0.9cm}<{\centering}p{0.9cm}<{\centering}
  }
  \toprule
  & Methods & Mean & Fam. & Fra. & Hor. & Lig. & M60 & Pan. & Pla. & Tra. \\
  \midrule
  \multirow{4}{*}{Supervised}
  & MVSNet \cite{yao2018mvsnet} &
  43.48 & 55.99 & 28.55 & 25.07 & 50.79 & 53.96 & 50.86 & 47.90 & 34.69 \\
  & CasMVSNet \cite{gu2020cascade} &
  \textbf{56.42} & 76.36 & \textbf{58.45} & \textbf{46.20} & \textbf{55.53} & 56.11 & 54.02 & \textbf{58.17} & 46.56 \\
  & CVP-MVSNet \cite{yang2020cost} &
  54.03 & \textbf{76.50} & 47.74 & 36.34 & 55.12 & \textbf{57.28} & \textbf{54.28} & 57.43 & 47.54 \\
  & PatchmatchNet \cite{wang2021patchmatchnet} &
  53.15 & 66.99 & 52.64 & 43.24 & 54.87 & 52.87 & 49.54 & 54.21 & \textbf{50.81} \\
  \midrule
  % \multirow{2}{*}{Two-Step}
  % & Yang et al. \cite{yang2021self} &
  % 46.71 & 64.95 & 38.79 & 24.98 & 49.73 & 52.57 & 51.53 & 50.66 & 40.45 \\
  % & U-MVSNet \cite{xu2021self} &
  % 57.15 & 76.49 & 60.04 & 49.20 & 55.52 & 55.33 & 51.22 & 56.77 & 52.63 \\
  % \midrule
  \multirow{4}{*}{Unsupervised}
  & MVS2 \cite{dai2019mvs2} &
  37.21 & 47.74 & 21.55 & 19.50 & 44.54 & 44.86 & 46.32 & 43.38 & 29.72 \\
  & M\^{}3VSNet \cite{huang2021m3vsnet} &
  37.67 & 47.74 & 24.38 & 18.74 & 44.42 & 43.45 & 44.95 & 47.39 & 30.31 \\
  & JDACS-MS \cite{xu2021self} &
  45.48 & 66.62 & 38.25 & 36.11 & 46.12 & 46.66 & 45.25 & 47.69 & 37.16 \\
  & Ours & \textbf{54.76} & \textbf{74.99} & \textbf{59.78} & \textbf{42.15} & \textbf{53.66} & \textbf{53.52} & \textbf{52.57} & \textbf{55.38} & \textbf{46.03} \\
  \bottomrule
\end{tabular}
  \label{tab:4_results_tanks}
\end{table*}

Tab. \ref{tab:4_effective_dtu} shows the comparison of GPU memory and runtime with our backbone method \cite{gu2020cascade}. As shown, benefit from probability volume splattering, our network only has a limited increase for both memory and runtime.

\subsection{Evaluation on Tanks and Temples dataset}

Furthermore, we evaluate the generalization ability of our DS-MVSNet method by using the model trained on the DTU dataset without fine-tuning process and reconstructing point clouds on the intermediate Tanks and Temples dataset. The input image size is $1920 \times 1056$ and the number of views N is 7. We use the camera parameters provided by MVSNet \cite{yao2018mvsnet} as the input and use the F-score \cite{knapitsch2017tanks} as the evaluation metric to measure the accuracy and completeness of the Tanks and Temples dataset.

\begin{figure*}[ht]
  \includegraphics[width=0.80\textwidth]{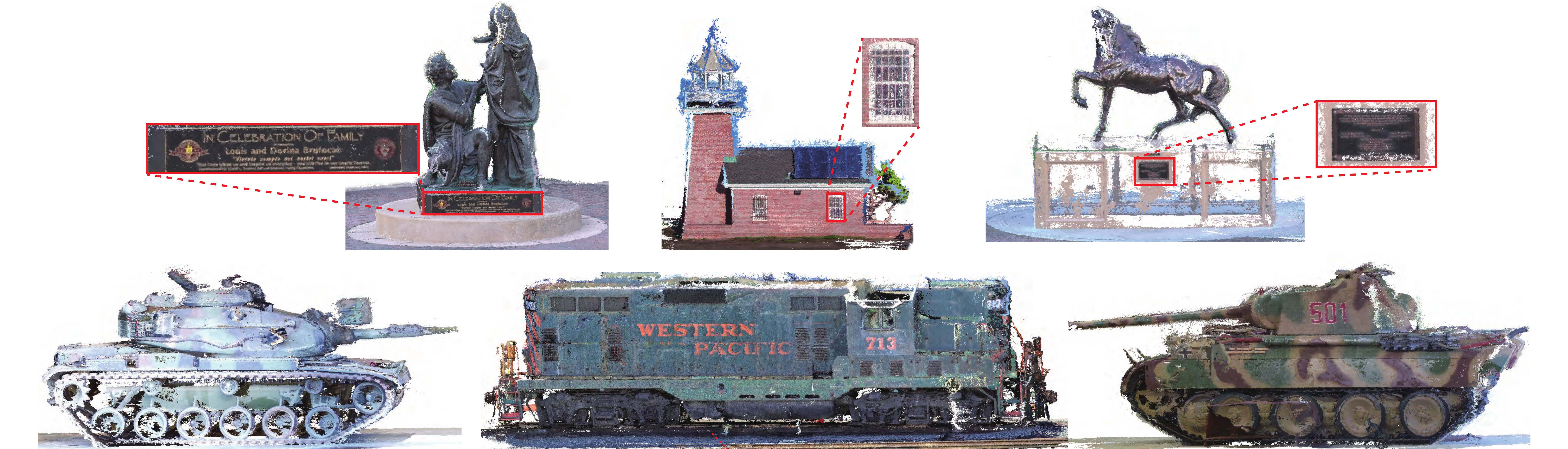}
  \caption{Visualization of the reconstructed 3D model on the intermediate partition of Tanks and Temples benchmark.}
  \label{fig:4_results_tanks}
\end{figure*}

We compare our method to state-of-the-art supervised and end-to-end unsupervised methods. The corresponding quantitative results on intermediate are reported in Tab. \ref{tab:4_results_tanks}.
Our method achieves state-of-the-art performance among all existing MVS methods and yields first place in all scenes, which fully confirms the generalization ability of our method. Our anonymous evaluation on the official leaderboard of \textit{Tanks and Temples} benchmark is named as DS-MVSNet.
Qualitative results of our point cloud reconstructions are shown in Fig. \ref{fig:4_results_tanks}.

\subsection{Ablation study}

Here we provide ablation studies analysis by evaluating the contribution of each part in our DS-MVSNet for the final reconstruction quality. 
Unless specified, all the following studies are done on DTU evaluation dataset.

\begin{table}[t]
  \caption{Ablation study of different components in our module on DTU evaluation set.
  RS: reference image smoothness. AS: adaptive depth hypotheses sampling.}
  \centering
  % \begin{tabular}{
%   p{2.5cm}
%   p{1.6cm}<{\centering}
%   p{1.6cm}<{\centering}
%   p{2cm}<{\centering}p{2cm}<{\centering}
%   p{1.5cm}<{\centering}p{1.5cm}<{\centering}p{1.5cm}<{\centering}
% }
%   \toprule
%   \multirow{2}{*}{Method} &
%   Num &
%   Ref &
%   \multicolumn{2}{c}{Adaptive Sampling} &
%   \multirow{2}{*}{Acc.} & \multirow{2}{*}{Comp.} & \multirow{2}{*}{Overall} \\
%   \cmidrule{4-5}
%   & Views & Smoothness & Ada. Bins & Ada. Gaussian & & & \\
%   \midrule
%   Baseline & 5 & & & & 0.424 & 0.383 & 0.403 \\
%   Baseline & 7 & & & & 0.428 & 0.350 & 0.389 \\
%   Baseline+RS & 7 & $\checkmark$ & & & 0.401 & 0.363 & 0.382 \\
%   Baseline+AB & 7 & & $\checkmark$ & & 0.412 & 0.354 & 0.383 \\
%   Baseline+AG & 7 & & & $\checkmark$ & 0.384 & 0.370 & 0.377 \\
%   Baseline+AB+AG & 7 & & $\checkmark$ & $\checkmark$ & 0.382 & 0.348 & 0.365 \\
%   Baseline+RS+AB+AG & 7 & $\checkmark$ & $\checkmark$ & $\checkmark$ & \textbf{0.374} & \textbf{0.347} & \textbf{0.361} \\
%   \bottomrule
% \end{tabular}

\begin{tabular}{
  p{1.6cm}
  p{1.0cm}<{\centering}
  p{0.3cm}<{\centering}
  p{0.3cm}<{\centering}
  p{0.8cm}<{\centering}p{0.8cm}<{\centering}p{0.8cm}<{\centering}
}
  \toprule
  Method & Views & RS & AS & Acc. & Comp. & Overall \\
  \midrule
  Baseline & 5 & & & 0.424 & 0.383 & 0.403 \\
  Baseline & 7 & & & 0.428 & 0.350 & 0.389 \\
  Baseline+RS & 7 & $\checkmark$ & & 0.401 & 0.363 & 0.382 \\
  Baseline+AS & 7 & & $\checkmark$ & 0.412 & 0.354 & 0.383 \\
  Baseline+RS+AS & 7 & $\checkmark$ & $\checkmark$ & \textbf{0.374} & \textbf{0.347} & \textbf{0.361} \\
  \bottomrule
\end{tabular}
  \label{tab:4_ablation_modules}
\end{table}

\textbf{Effective of Different Modules Components}
Tab. \ref{tab:4_ablation_modules} shows the improved reconstruction performances through the number of views, the reference image smoothness module, and the adaptive sampling.
Observed from Tab. \ref{tab:4_ablation_modules}, more different views can improve the reconstruction result. So for other analyses, the number of views is all set to 7.
Besides, the reference image smoothness module helps to reconstruct more accurate points by weakening the effect of shadows, mentioned in Sec. \ref{subsec:training_loss}.
And the completeness of reconstruction results improves obviously under the adaptive sampling module, which can provide more accurate depth hypotheses.

\textbf{Effective of Adaptive Depth Hypotheses Sampling.} To quantitatively measure the effectiveness of the manner in depth sampling, five experiments are executed as shown in Tab. \ref{tab:4_ablation_samples}.
In the method Comb-A, we used uniform sampling for all stages.
Then, for the first stage, we replaced it with adaptive bins sampling. As shown in Comb-A and Comb-B, we achieved better results than the uniform sampling. Because adaptive gaussian sampling depends on the initial value, so we only compared uniform and adaptive bins sampling in the first stage. 
For the second stage, we compared uniform, adaptive bins, and adaptive Gaussian sampling. Illustrated in Comb-(B, C, D), adaptive bins and gaussian sampling can achieve comparable results, but adaptive gaussian is a better choice because of no learnable parameters.
However, the adaptive bins sampling cannot be trained in the third stage since the 24GB GPU memory can't meet its memory overhead with increasing resolution, and it also cannot be used even on the 32GB V100s GPU.
Thus, we only compared uniform and adaptive gaussian sampling in the third stage, and Comb-(D, E) shows that adaptive Gaussian sampling achieves a performance improvement.
%The adaptive bins sampling can not be trained in the third stage on our machine, because of no enough GPU memory with the resolution increases. So we only compared uniform and adaptive gaussian sampling in the third stage, Comb-(D,E) show the adaptive gaussian sampling could achieve a performance improvement.

\begin{table}[t]
  \caption{The quantitative result with different sampling strategies in each stage. UniS: uniform sampling. AdaB: adaptive bins. AdaG: adaptive gaussian.}
  \centering
  \begin{tabular}{
  p{1.2cm}<{\centering}
  p{0.8cm}<{\centering}
  p{0.8cm}<{\centering}
  p{0.8cm}<{\centering}
  p{0.8cm}<{\centering}p{0.8cm}<{\centering}p{0.8cm}<{\centering}
}
  \toprule
  Method & Stage1 & Stage2 & Stage3 & Acc. & Comp. & Overall \\
  \midrule
  Comb-A & UniS & UniS & UniS & 0.401 & 0.363 & 0.382 \\
  Comb-B & AdaB & UniS & UniS & 0.380 & 0.359 & 0.370 \\
  Comb-C & AdaB & AdaB & UniS & 0.376 & 0.354 & 0.365 \\
  Comb-D & AdaB & AdaG & UniS & 0.377 & 0.351 & 0.364 \\
  Comb-E & AdaB & AdaG & AdaG & \textbf{0.374} & \textbf{0.347} & \textbf{0.361} \\
  \bottomrule
\end{tabular}
  \label{tab:4_ablation_samples}
\end{table}

\textbf{Effective of Different Loss components}
In Tab. \ref{tab:4_ablation_loss}, we discuss the influence of different losses in the network for the final geometry reconstruction.
Here, five experiments are executed to study the contribution of the reference match cost loss, source match cost loss, depth smoothness loss, and depth consistency loss. There are two components in the cost loss, including photo consistency loss and ssim loss.
In our network, we generate the depth and image of the source view through the reference depth hypotheses, thus we add the reference match cost loss compared to other networks.
As shown in Tab. \ref{tab:4_ablation_loss}, we can easily observe that the reference match cost loss plays an important role in the approach compared with source match cost loss.
Based on reference match cost, source match cost loss can provide additional supervision to help generate more points.
Besides, depth smoothness components can effectively improve the performance on completeness metrics, and depth consistency loss achieves a good balance between accuracy and completeness.

\begin{table}[t]
  \caption{Ablation study of different components of our proposed loss. Lower means better. $\mathcal{L}_{mc}$: The sum of Photometric Consistency and SSIM Loss. $\mathcal{L}_{ds}$: Depth Smoothness Loss. $\mathcal{L}_{dc}$: Depth Consistency Loss.
  }
  \centering
  \begin{tabular}{
  p{0.8cm}<{\centering}p{0.8cm}<{\centering}p{0.8cm}<{\centering}p{0.8cm}<{\centering}
  p{0.8cm}<{\centering}p{0.8cm}<{\centering}p{0.8cm}<{\centering}
}
  \toprule
  \multicolumn{2}{c}{Match Cost} &
  \multirow{2}{*}{$\mathcal{L}_{ds}$} & \multirow{2}{*}{$\mathcal{L}_{dc}$} &
  \multirow{2}{*}{Acc.} & \multirow{2}{*}{Comp.} & \multirow{2}{*}{Overall} \\
  \cmidrule{1-2}
%   $\mathcal{L}_{pc}^{ref} + \mathcal{L}_{ssim}^{ref}$ &
%   $\mathcal{L}_{pc}^{src} + \mathcal{L}_{ssim}^{src}$ & & & & & \\
  $\mathcal{L}_{mc}^{ref}$ &
  $\mathcal{L}_{mc}^{src}$ & & & & & \\
  \midrule
  \checkmark & & & & 0.417 & 0.387 & 0.402 \\
  & \checkmark & & & 0.433 & 0.396 & 0.415 \\
  \checkmark & \checkmark & & & 0.395 & 0.379 & 0.387 \\
  \checkmark & \checkmark & \checkmark & & 0.391 & \textbf{0.342} & 0.367 \\
  \checkmark & \checkmark & \checkmark & \checkmark & \textbf{0.374} & 0.347& \textbf{0.361} \\
  \bottomrule
\end{tabular}
  \label{tab:4_ablation_loss}
\end{table}

\section{Conclusion}

In this paper, we propose an end-to-end unsupervised MVS structure with the source depths synthesis. Compared with previous unsupervised methods, our proposed method can simplify the training process benefits from inherent information of probability volume. For example, we don't need to generate the pseudo depth labels, and not require data augmentation, processed optical flow and pre-trained network. The DS-MVSNet achieves competitive  performance compared with unsupervised networks, and we have shown this through the state-of-the-art performance on DTU and Tanks and Temples benchmarks.
One current limitation of our approach is handling the occlusion between different views as the depth synthesis and image rendering fail in these areas.
Besides, the difference in the multiple views is limited for the texture-less areas, thus it's still a challenge in MVS task, as the Horse scene in Fig. \ref{fig:4_results_tanks}.
We will address these in the future.

\bibliographystyle{ACM-Reference-Format}
\bibliography{main}

\end{document}